\documentclass[letterpaper]{article} 
\usepackage[]{aaai25}  
\usepackage{times}  
\usepackage{helvet}  
\usepackage{courier}  
\usepackage[hyphens]{url}  
\usepackage{graphicx} 
\usepackage{natbib}  
\usepackage{caption} 
\usepackage{algorithm}
\usepackage{algorithmic}

\usepackage{multirow}
\usepackage{amssymb}
\usepackage{amsmath}
\usepackage{cleveref}

\usepackage{newfloat}
\usepackage{listings}
\DeclareCaptionStyle{ruled}{labelfont=normalfont,labelsep=colon,strut=off} 
\floatstyle{ruled}
\newfloat{listing}{tb}{lst}{}
\floatname{listing}{Listing}
\nocopyright

\title{OccLLaMA: An Occupancy-Language-Action Generative World Model for Autonomous Driving}
\author{
    Julong Wei\textsuperscript{\rm 1}, 
    Shanshuai Yuan\textsuperscript{\rm 1}, 
    Pengfei Li\textsuperscript{\rm 2}, 
    Qingda Hu\textsuperscript{\rm 1}, 
    Zhongxue Gan\textsuperscript{\rm 1}\equalcontrib, 
    Wenchao Ding\textsuperscript{\rm 1}\equalcontrib
}
\affiliations{
    \textsuperscript{\rm 1}Academy for Engineering \& Technology, Fudan University \\ 
    \textsuperscript{\rm 2}Institute for AI Industry Research, Tsinghua University \\
    \{jlwei23, ssyuan23\}@m.fudan.edu.cn, li-pf22@mails.tsinghua.edu.cn, qdhu24@m.fudan.edu.cn, \{ganzhongxue, dingwenchao\}@fudan.edu.cn
}

\usepackage{bibentry}

\begin{document}

\maketitle

\begin{abstract}

The rise of multi-modal large language models(MLLMs) has spurred their applications in autonomous driving. Recent MLLM-based methods perform action by learning a direct mapping from perception to action, neglecting the dynamics of the world and the relations between action and world dynamics. In contrast, human beings possess world model that enables them to simulate the future states based on 3D internal visual representation and plan actions accordingly. To this end, we propose OccLLaMA, an occupancy-language-action generative world model, which uses semantic occupancy as a general visual representation and unifies vision-language-action(VLA) modalities through an autoregressive model. Specifically, we introduce a novel VQVAE-like scene tokenizer to efficiently discretize and reconstruct semantic occupancy scenes, considering its sparsity and classes imbalance. Then, we build a unified multi-modal vocabulary for vision, language and action. Furthermore, we enhance LLM, specifically LLaMA, to perform the next token/scene prediction on the unified vocabulary to complete multiple tasks in autonomous driving. Extensive experiments demonstrate that OccLLaMA achieves competitive performance across multiple tasks, including 4D occupancy forecasting, motion planning, and visual question answering, showcasing its potential as a foundation model in autonomous driving.

\end{abstract}

%


\section{Introduction}

Rencent years, we have witnessed a significant breakthrough in multi-modal large language models(MLLMs) capable of integrating various modalities, such as language, image, audio, which has accelerated the development of embodied artificial intelligence (Embodied AI). Nevertheless, a general agent, which can address multiple tasks in real-world has yet to emerge. This is inherently because existing MLLMs perform action by learning a direct mapping from perception to action, neglecting the dynamics of the world and the relations between action and world dynamics. In contrast, human beings possess world model that enables them to simulate the future states based on 3D internal visual representation and plan actions accordingly. Therefore, exploring how to construct the agent's world model is crucial for the advancement of embodied intelligence.

Autonomous driving, as a representative application of Embodied AI, has witnessed extensive research on world models. However, \textit{the precise definition of a world model for autonomous driving remains an open question.} Current world models for autonomous driving focus on sensor prediction tasks such as video prediction~\cite{wang2024driving}, point cloud prediction~\cite{zhang2023learning} and occupancy prediction~\cite{11}. Yet they fail to simultaneously achieve forecasting scene evolution, language reasoning, and interaction with the real world. Thus, we propose that a model capable of unifying the modeling of vision, language, and actions(VLA), akin to human abilities, would be a promising candidate for an autonomous driving world model.

However, two challenges are crucial and need to be solved for building a VLA world model. The first is to build a general 3D visual representation that facilitates both understanding and generation, and the second is to develop a multi-modal framework capable of accommodating VLA modalities. Recent years, semantic occupancy (Occ) has gained significant attention as a general 3D visual representation. It can describe fine-grained 3D structure, while also containing high-level semantic information, making it well-suited for aligning space and semantics. Meanwhile, the feasibility of vision generation based on autoregressive language models has been thoroughly validated, with performance comparable to diffusion models, which are specialists in visual generation. These provide valuable insights for addressing the challenges and constructing a VLA world model based on an autoregressive model with Occ visual representation.

\begin{figure*}[!h]
    \centering
    \includegraphics[width=\linewidth]{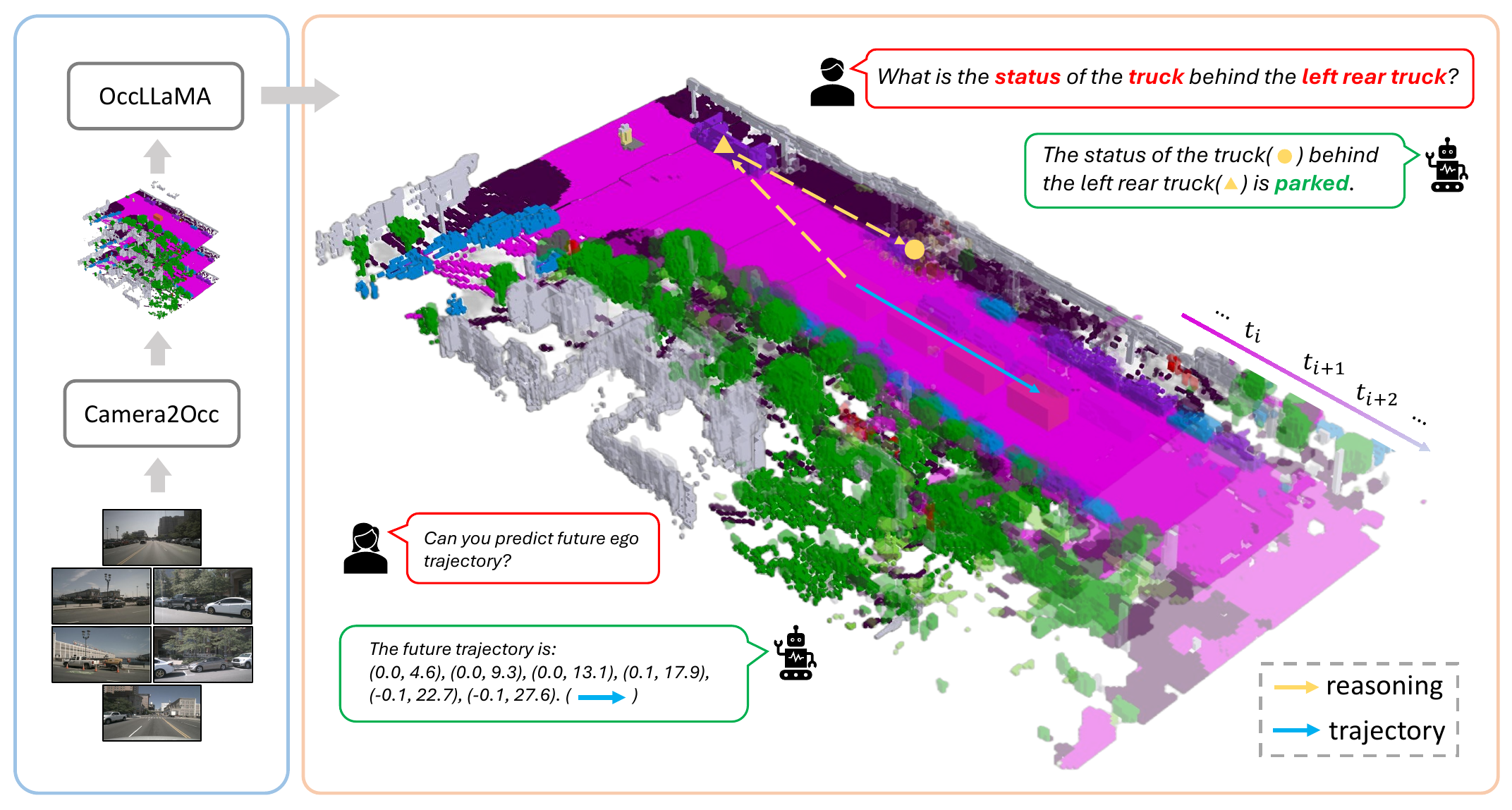}
    \caption{OccLLaMA accepts occupancy data from existing occupancy prediction algorithms, and performs a series of tasks, including scene understanding and reasoning, 4D occupancy prediction, and motion planning.}
    \label{fig:multitask}
\end{figure*}

Based on the above observations, we propose OccLLaMA, a unified 3D occupancy-language-action generative world model, which unifies VLA-related tasks including but not limited to scene understanding, planning, and 4D occupancy forecasting, as shown in \Cref{fig:multitask}. To enable OccLLaMA with the ability to understand and generate vision modality, we choose Occ as a general visual representation and introduce a novel scene tokenizer to construct discrete scene vocabulary effectively, considering sparsity and classes imbalance. Then, by combining scene vocabulary, language vocabulary, and action vocabulary, we construct a unified multi-modal vocabulary for VLA-tasks, which provides a foundation for integrating VLA in one model. Furthermore, we enhance LLM, specifically LLaMA~\cite{touvron2023llama}, to implement next token/scene prediction on the unified multi-modal vocabulary, building a VLA world model similar to humans.


We summarize our contributions as follows:
\begin{itemize}

\item An occupancy-language-action generative world model, OccLLaMA, which uses Occ as visual representation and involves multiple tasks through unified multi-modal vocabulary and enhanced autoregressive model based on LLaMA.

\item A novel scene tokenizer that efficiently discretize and reconstruct Occ scenes, considering sparsity and classes imbalance.

\item Extensive experiments compared to SOTA methods, which achieves the competitive performance across multiple tasks, including 4D occupancy forecasting, motion planning, and visual question answering.
\end{itemize}

\section{Related Works}

\subsection{MLLMs in Autonomous Driving}

The advancement of LLMs explore new paradigms in autonomous driving, including scene understanding~\cite{LP,HILMD,CAN,MDTGPT}, end-to-end decision-making~\cite{DRIVELM,19,21}. 
LLM-driven decision-making methods have shown potential in addressing interpretability and generalization challenges in learning-based systems by making inferences in text space. For real-world autonomous driving, various techniques have emerged to convey environmental information to models, with research advancing to extend input modalities more effectively.
It includes template-based scene description in natural language~\cite{13,14,15}, vector embedding input combined with language prompts~\cite{16}, camera-perceived image embedding~\cite{17,18},etc. It remains to be validated whether more modalities mean high performance. For example, point clouds do not show the ability to enhance performance in DriveMLM~\cite{19}. Moreover, the output modality in past work tends to be relatively homogeneous, which somewhat limits the model's accuracy and stability of decision making and closed-loop feedback capability. Thus there is an opportunity for multimodal large language models(MLLM) and world models(WM) to shake hands.

\begin{figure*}[!h]
    \centering
    \includegraphics[width=\linewidth]{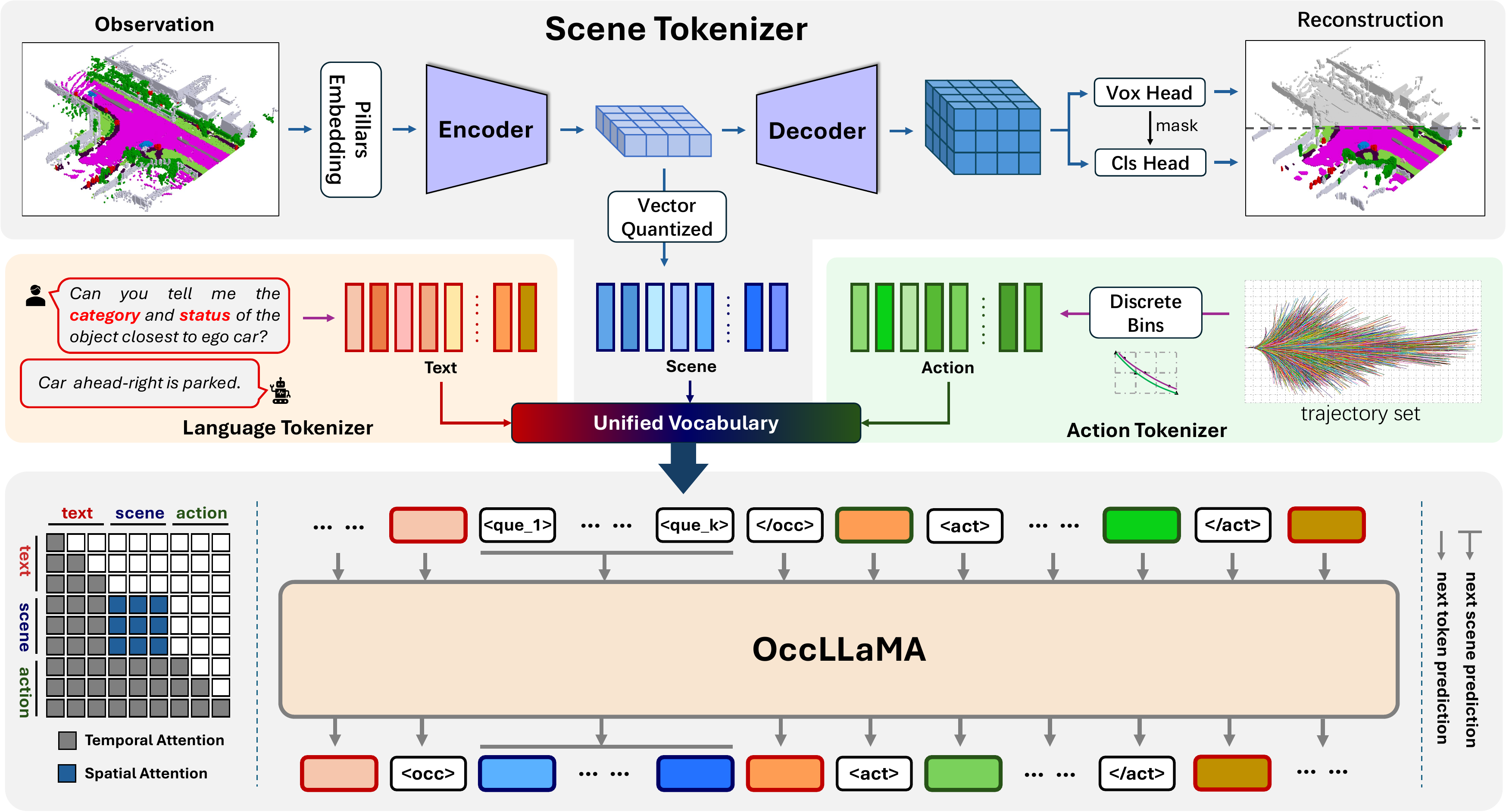}
    \caption{Overview of the OccLLaMA architecture. The core components are the Scene Tokenizer and the Generative World Model. The Scene Tokenizer discretizes the Occ representation into scene vocabulary, which are then combined with the language and action vocabularies to construct a unified vocabulary. Furthermore, the  Generative World Model performs next token / scene prediction on the unified vocabulary space including scene understanding, reasoning and action.}
    \label{fig:pipeline}
\end{figure*}

\subsection{World Model in Autonomous Driving}

World Models aim to predict future scene based on the action and observations~\cite{1}.In autonomous driving, world models are often used for data generation and decision making. Various models represents the scene in different spaces, they can be divided into 2D image representation~\cite{2, 3, 4, 5, 6}, 3D point clouds representation~\cite{7,8,9} and 3D occupancy representation~\cite{10,11,11.5}. 
Visual world models using 2D image representations offer scalability due to sensor flexibility but lack 3D scene comprehension. While 3D point clouds representations address this problem, they lack semantic information.
Some works~\cite{4,10} focus on multi-modal representation, but it is difficult to align the features of results generated in different modality. Therefore, integrating the 3D scene representation and semantic understanding is a promising way to model the scene evolution. Unlike the paradigm of ~\cite{10}generating the 3D occupancy without semantics meaning attached and scene representation in camera and point cloud form separately. Refer to the paradigm of Occupancy World~\cite{11},representation on 3D occupancy space with semantics is adopted in this work.

\subsection{Autoregressive Visual Generation}

Autoregressive (AR) Visual Generation refers to models use autoregressive methods to generate images. Early models such as VQVAE~\cite{22}, VQGAN~\cite{23} and Dalle~\cite{DALLE}, convert images into discrete tokens and generate them sequentially faced limitations in output performance, scalability. Then diffusion models~\cite{24,25,26}dominated the field of visual generation due to distinct paradigms. 
Recently, the simplicity of the autoregressive model has enabled uniform understanding and generation, effectively scaling up big data, leading to notable success and increased attention.
VAR~\cite{27} model enables GPT-based autoregressive models to surpass diffusion models in image generation. Llama-Gen~\cite{29} outperforms diffusion models in conditional image generation, showing that pure autoregressive models can serve as a foundation for image generation without visual signal inductive bias. Integrating AR language models with vision generation remains challenging, particularly in creating unified models for both language and vision tasks.

\section{Method}
\subsection{Overview}


We propose OccLLaMA, a uniform occupancy-language-action framework. As illustrated in \Cref{fig:pipeline}, the core components of OccLLaMA include the scene tokenizer (\Cref{sec:tokenizer}) and the occupancy-language-action generative world model (\Cref{sec:worldmodel}). To involve multitask, we introduce a three-stage training scheme (\Cref{sec:trainstage}) for the training of scene tokenizer, occupancy-language-action pre-training, and instruction tuning.

\subsection{Scene Tokenizer} 
\label{sec:tokenizer}
To represent scenes using discrete tokens, a common approach is to employ a VQVAE-like architecture~\cite{van2017neural}. However, approximately 90\% of the grids~\cite{occ3d} in occupancy are filled with air, leading to significant sparsity. 
Existing methods that apply dense convolutional operators to the air category are both  expensive and inefficient. Additionally, the imbalance between categories further hinders learning efficiency. To address these challenges, we introduce a sparse encoding strategy for the encoder, inspired by point cloud processing techniques. Meanwhile, we decouple the non-occupied category from other semantic categories, allowing for more efficient scene reconstruction.

\subsubsection{Encoder} The original scene is represented as $x \in \mathbb{R}^{H \times W \times D}$, where the 3D space is divided into dense $H \times W \times D$ voxels, and each voxel is assigned a semantic label $l \in \mathbb{R} $. We then sparsify $x$ into $y \in P^{H \times W}$ by discarding the air voxels and representing the semantic occupancy voxels as a 1D pseudo-point cloud set $P=\{{p_i}\}_{i=1}^N$ arranged along the BEV direction, where $N$ is the number of non-air voxels within the current pillar. Each point $p_i$ is a vector $(d,l)$ with height $d$ and semantic label $l$. We then aggregate the pseudo-point cloud features using a pillars embedding~\cite{pointpillars} and employ a swin-transformer block~\cite{swin} to obtain the BEV feature map $z = E(y) \in \mathbb{R}^{\frac{H}{r} \times \frac{W}{r} \times c}$, where $r$ is downsampling rate, and $c$ is the latent feature dimension.

\subsubsection{Quantification} To obtain discrete representations, we next transform $z$ into a collection of codebook entries $\hat{z}$ through vector quantization. The learnable codebook $Z = \{\hat{z}_i\}^K_{i=1}$ consists of $K$ vectors, each with a dimension of $c$. The quantization process $Q(\cdot)$ replaces each $z_i$ with its nearest codebook entry $\hat{z}_k$ in $Z$, expressed as:
\begin{align}
\hat{z}_i=Q(z_i):=\arg\min_{\hat{z}_k\in Z}\left\| z_i - \hat{z}_k \right\|_2
\end{align}
\subsubsection{Decoder} Due to the loss of height information in the BEV feature map after quantization, the decoder restores dense 3D voxel features by stacking convolution block and up-sampling layer. Specifically, to address classes imbalance, we instantiate lightweight voxel head and class head separately to decode geometric and semantic information of occupancy. Notably, the voxel head provides an occupied mask for the class head, allowing us to supervise the semantics of the occupied voxels only.
\subsubsection{Loss}
To train this scene tokenizer, we follow OccWorld~\cite{11} to utilize three loss functions for optimizing, where cross-entropy loss $\mathcal{L}_c$ and lovasz-softmax loss $ \mathcal{L}_l$ for geometry $ge$ and semantics $se$ reconstruction learning, and the embedding loss $\mathcal{L}_e$ for codebook learning.
\begin{equation}
\mathcal{L} = \lambda_1 \mathcal{L}_c^{ge} + \lambda_2\mathcal{L}_l^{ge} +
\lambda_3 \mathcal{L}_c^{se} + \lambda_4\mathcal{L}_l^{se}
+ \lambda_5\mathcal{L}_e 
\label{align:loss}
\end{equation}

\subsection{Generative World Model}
\label{sec:worldmodel}

\subsubsection{Unified Vocabulary} 
Employing scene tokenizer in \Cref{sec:tokenizer}, an occupancy scene $x$ can be mapped and flattened into a sequence $\hat{z}^{1:L} \in \mathbb{R}^c$, where $ L = \frac{H}{r} \times \frac{W}{r}$, allowing for joint representation with similar language vocabulary $V_t = \{ v_t^i\}_{i=1}^{K_t}$ in original LLMs. Specifically, we first represent scene tokens $\hat{z}^{1:L}$ as a sequence of indices $s^{1:L} = \{s^i\}_{i=1}^L$, where $s^i$ corresponds to the code index number of scene tokens $\hat{z}^{1:L}$. So we can build a scene vocabulary $V_s = \{ v_s^i\}_{i=1}^{K_s}$, which is order-preserving to our scene codebook $Z$. Since it is non-trivial to output fine-grained numerical results using general LLMs, we divide the coordinates of waypoints into $N$ bins empirically based on the statistics of the trajectory set and map waypoint to the nearest bin, to build an action vocabulary $V_a = \{ v_a^i\}_{i=1}^{K_a}$. Additionally, we add several special functional tokens $ \{ v_f^i\}_{i=1}^{K_f}$, such as \texttt{<occ>}, \texttt{</occ>}, \texttt{<act>}, \texttt{</act>} to denote modality boundaries; \texttt{<que\_i>} to assist in next scene prediction. Thus, we can build a unified occupancy-language-action vocabulary $V = \{ V_s, V_t, V_a, \{ v_f^i\}_{i=1}^{K_f} \}$ to formulate diverse tasks in a generate format, where both input and output can be one of the three modalities or a mixture, depending on the specific task to be solved.

\subsubsection{Next Token / Scene Prediction} 
We observe that both language and action are temporal sequences, making the tokens within these sequences naturally suitable for temporal attention with original causal masks and next token prediction mechanisms. However, the tokens within a scene sequence do not inherently follow a temporal order, and the sequence length tend to greater than language and action. If next token prediction is performed line by line within a scene, it fails to capture the spatial relationships and incurs high computational costs. To address these issues, we introduce a next scene prediction while preserving the next token prediction.

As illustrated in \Cref{fig:pipeline}, we implement spatial attention at the positions corresponding to scene tokens to better capture the spatial relationships within the scene. Correspondingly, we initialize learnable scene queries to predict the entire scene in one forward step, enabling better interaction among tokens within the scene and significantly reducing inference time. In \Cref{alg:autoregressive}, we provide a detailed explanation of the mechanism for executing next token / scene prediction simultaneously.

\begin{algorithm}[tb]
\caption{Next Token / Scene Prediction}
\label{alg:autoregressive}
\textbf{Input}: question sequence $x = \{ x_i\} \in V$ \\
\textbf{Parameters}: max length $L$, scene length $S$ \\
\textbf{Output}: completed sequence $x$
\begin{algorithmic}[0] 
\STATE Init generative world model as $M$
\STATE $q = [\texttt{<que\_i>} \text{ for } i \text{ in range}(S)]$ 
\WHILE{$x[-1] \neq \texttt{<eot\_id>}$ \textbf{and} $\text{len}(x) < L$}
    \IF{$x[-1] == \texttt{<occ>}$}
        \STATE  $x.\text{append}(q)$ 
        \STATE $x[-K:] = M(x)[-K:]$
        \STATE $x.\text{append}(\texttt{</occ>})$
    
    \ELSE
        \STATE $x.\text{append}(M(x)[-1])$
    \ENDIF
\ENDWHILE
\STATE \textbf{return} $x$
\end{algorithmic}
\end{algorithm}

\begin{table*}[!h]
\centering
\begin{tabular}{l|c|ccccc|ccccc}
\cline{1-12}
\multirow{2}{*}{Method} & \multirow{2}{*}{Input} & \multicolumn{5}{c|}{MIOU(\%)$\uparrow$}   & \multicolumn{5}{c}{IOU(\%)$\uparrow$}    \\
  &  & Recon. & 1s & 2s & 3s & Avg. & Recon. & 1s & 2s & 3s & Avg. \\ 
\cline{1-12}
OccWorld-F & Camera &  20.09  &  8.03  & 6.91   &  3.54  &  6.16  &  35.61   &     23.62   &  18.13  &  15.22  &  18.99    \\
OccWorld-O & Occ    &   66.38     &  \textbf{25.78}  &  15.14  &  10.51  &  17.14   &   62.29     &  \textbf{34.63}  &  25.07  &  20.18  &  26.63            \\
\textbf{Ours-F}  & Camera   &   37.38     &  10.34  &  8.66  &  6.98  &   8.66  &   38.92        &  25.81  &  23.19  &   19.97 &   22.99  \\
\textbf{Ours-O}  & Occ      &   \textbf{75.20}     &  25.05  &  \textbf{19.49}  &  \textbf{15.26}  &  \textbf{19.93}   &   \textbf{63.76}     &  34.56  &  \textbf{28.53}  &  \textbf{24.41}  &  \textbf{29.17} \\
\cline{1-12}
\end{tabular}
\caption{Quantitative results of 4D occupancy forecasting. Recon. refers to reconstruction performance of scene tokenizer.}
\label{tab:forecasting}
\end{table*}

\subsection{Train Stage}
\label{sec:trainstage}
Our training scheme includes three stages:

\subsubsection{Training of scene tokenizer} We first focus on learning the scene codebook to represent occupancy as discrete tokens, using the objective defined in \Cref{align:loss}. Once optimized, the scene tokenizer remains unchanged throughout the subsequent stages of the pipeline.

\subsubsection{3D Occupancy-Language-Action pre-training} We focus on aligning occupancy-language-action modalities in this stage. We use world model objective and scene-caption objective for full parameter pre-training, the former supervises the alignment between occupancy and action to learn the evolution of the world, the latter supervises the alignment between occupancy and language to learn the semantic understanding of the 3D scene.

\subsubsection{Instruction tuning} In this stage, we fine-tunes the model based on prompt-based instructions for different scene understanding and planning tasks by LoRA~\cite{lora}.

\section{Experiments}
\subsection{Experiment Settings}

\subsubsection{Dataset} NuScenes~\cite{caesar2020nuscenes} is a widely recognized foundation dataset in autonomous driving. The dataset comprises 700 training videos and 150 validation videos, each spanning 20 seconds with a key frame rate of 2Hz. Occ3D~\cite{tian2024occ3d} is a large-scale dataset for 3D occupancy based on NuScenes, providing a semantic occupancy representation for each frame. NuScenes-QA~\cite{qian2024nuscenes} is a multi-modal visual question answering dataset based on Nuscenes. It encompasses five categories of questions: existence, counting, query-object, query-status, and comparison, which further categorized into zero-hop and one-hop by complexity. For aligning occupancy and language modalities, we collecte a large caption dataset based on NuScenes. Specifically, this dataset matches occupancy frames with the positions, classes, states, and future trajectories of objects appearing on them.

\subsubsection{Implementation Details} 
For most comparisons, we set language model backbone as LLaMA-3.1-8b and the scene tokenizer parameters as $50\times256\times2048$. For VQA comparison, we set language model backbone as LLaMA-2-7b and scene tokenizer resolution as $25\times25$ for fairness. We employ the AdamW optimizer for all the training. Scene tokenizer is trained with learning rate of \(10^{-4}\), batch size of $4$, $\lambda_1=\lambda_3 = 10$, $\lambda_2 = \lambda_4 = 5$, and $\lambda_5 = 5$, while Generative Model is trained with learning rate of \(10^{-4}\) and batch size of $1$ in pre-training strage, \(5 \times 10^{-5}\) and $4$ in instruction tuning stage. The Scene tokenizer undergoes 100 epoch on 8 RTX 4090 GPUs, while the Generative Model undergoes $10$ epoch in pre-training stage and $5$ epoch in instruction tuning stage on 8 V100 GPUs.

\begin{figure*}[h]
    \centering
    \includegraphics[width=\linewidth]{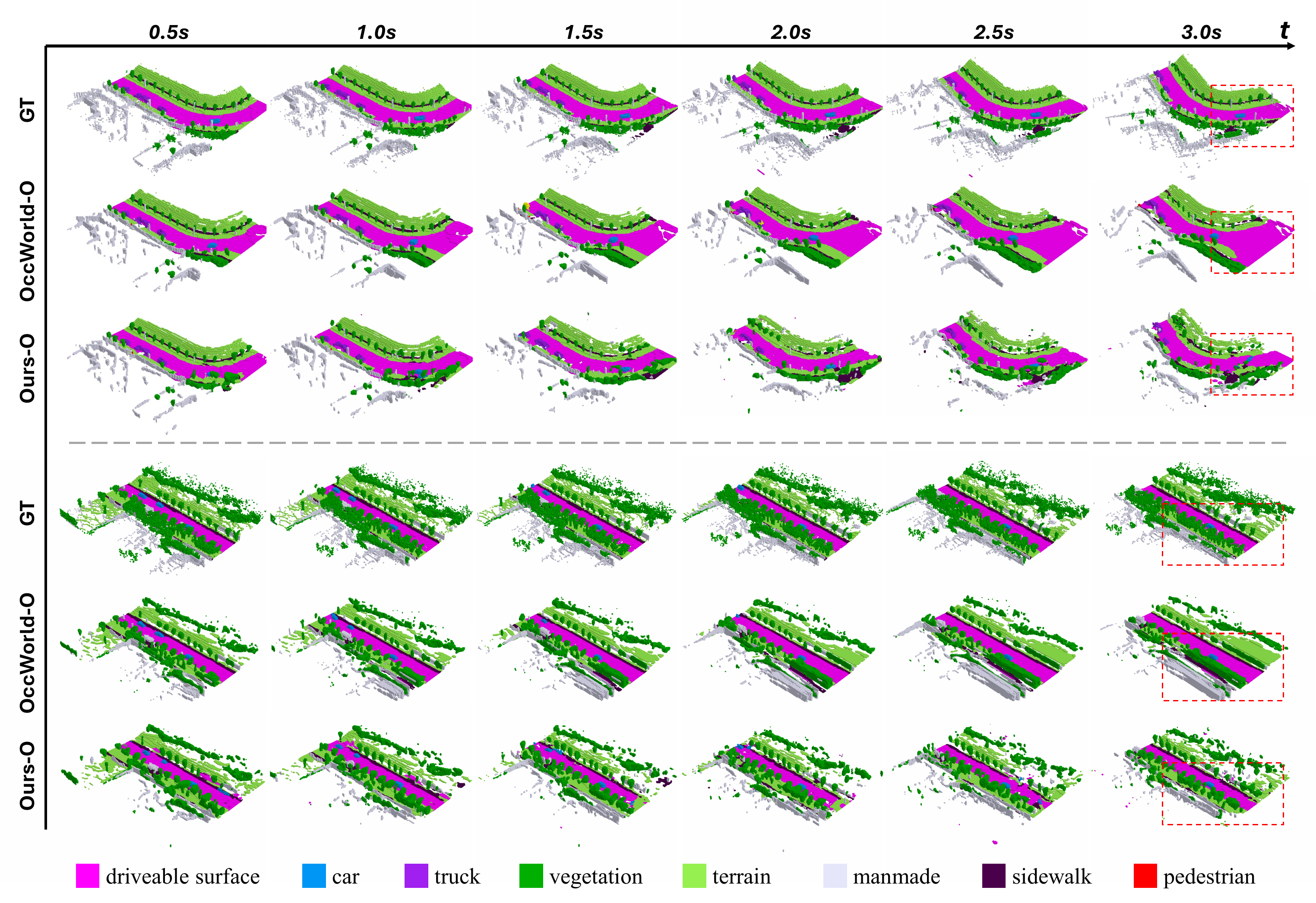}
    \caption{Qualitative results of 4D occupancy forecasting. Compared to OccWorld, OccLLaMA exhibits superior long-term temporal modeling capabilities, allowing it to maintain strong performance at the 3-second mark. In contrast, OccWorld leads to significant discrepancies from the ground truth.}
    \label{fig:vis_forecasting}
\end{figure*}


\subsection{Results and Analysis}
\subsubsection{4D Occupancy Forecasting} The task aims to forecast the future 3D occupancy scene given a few historical occupancy inputs. Specifically, we follow existing works~\cite{11} and used a 2-second historical frames to forecast the subsequent 3 seconds and use mIoU and IoU as the main evaluation metrics. As illustrated in \Cref{tab:forecasting}, we compare OccLLaMA with the state-of-the-art approach, OccWorld, in two settings: using ground-truth 3D occupancy (-O), using predicted results from FBOCC~\cite{li2023fb} based on camera data (-F). Firstly, we observe that our scene tokenizer demonstrates superior scene reconstruction capabilities. Additionally, OccLLaMA achieves a competitive forecasting result within 1s, and significantly outperformes OccWorld over longer time, highlighting its enhanced long-term prediction capabilities. Furthermore, OccLLaMA-F can be regarded as an end-to-end pipeline as it takes cameras as input. Despite the complexity of the task, OccLLaMA consistently exhibits strong predictive performance. We present visualizations consistent with the above conclusions in \Cref{fig:vis_forecasting}. 



\begin{table*}[!h]
\centering
\setlength{\tabcolsep}{1.4mm}
\begin{tabular}{l|cc|cccc|cccc}
\cline{1-11}
\multirow{2}{*}{Method} & \multirow{2}{*}{Input} & \multirow{2}{*}{Sup.} & \multicolumn{4}{c|}{L2(m)$\downarrow$} & \multicolumn{4}{c}{Coll.(\%)$\downarrow$} \\
& & & 1s & 2s & 3s & Avg. & 1s & 2s & 3s & Avg. \\ 
\cline{1-11}
IL & LiDAR & None & 0.44 & 1.15 & 2.47 & 1.35 & 0.08 & 0.27 & 1.95 & 0.77 \\
NMP & LiDAR & Box \& Motion & 0.53 & 1.25 & 2.67 & 1.48 & \underline{0.04} & \textbf{0.12} & \underline{0.87} & \underline{0.34} \\
FF & LiDAR & Freespace & 0.55 & 1.20 & 2.54 & 1.43 & 0.06 & \underline{0.17} & 1.07 & 0.43\\
\cline{1-11}
ST-P3 &Camera & Map \& Box \& Depth & 1.33 & 2.11 & 2.90 & 2.11 & 0.23 & 0.62 & 1.27 & 0.71 \\
UniAD &Camera & Map \& Box \& Motion \& Track \& Occ & 0.48 & \textbf{0.96} & \textbf{1.65} & \textbf{1.03} & 0.05 & \underline{0.17} & \textbf{0.71} & \textbf{0.31} \\
VAD &Camera & Map \& Box \& Motion & 0.54 & 1.15 & \underline{1.98} & 1.22 & \underline{0.04} & 0.39 & 1.17 & 0.53 \\
OccWorld-F & Camera & Occ & 0.45& 1.33 & 2.25&1.34 &0.08 &0.42 &1.71&0.73 \\
\textbf{Ours-F }& Camera& Occ & \underline{0.38} &1.07  &  2.15&  1.20  & 0.06  & 0.39 & 1.65 &   0.70 \\      
\cline{1-11}
OccNet &Occ & Map \& Box & 1.29 & 2.31 & 2.98 & 2.25 & 0.20 & 0.56 & 1.30 & 0.69 \\
OccWorld-O & Occ & None & 0.43 & 1.08 & 1.99 & 1.17 & 0.07 & 0.38 & 1.35 & 0.60 \\
\textbf{Ours-O$^\dag$ }& Occ& None &  \underline{0.38} & 1.06  & 2.08 & 1.18  &  \textbf{0.02}  &   \underline{0.17}  &  1.39 &    0.53 \\
\textbf{Ours-O} & Occ& None &  \textbf{0.37} & \underline{1.02}  & 2.03 & \underline{1.14}  &  \underline{0.04}  &   0.24  &  1.20 &    0.49 \\
\cline{1-11}
\end{tabular}
\caption{Quantitative results of motion planning. Ours-O$^\dag$ refers to output trajectories without scene predictions.}
\label{tab:planning}
\end{table*}

\begin{table*}[!h]
\centering
\setlength{\tabcolsep}{1.1mm}
\begin{tabular}{c|c|ccc|ccc|ccc|ccc|ccc|c}
\hline
\multirow{2}{*}{Method} & \multirow{2}{*}{input} & \multicolumn{3}{c|}{exist(\%)$\uparrow$} & \multicolumn{3}{c|}{count(\%)$\uparrow$} & \multicolumn{3}{c|}{object(\%)$\uparrow$} & \multicolumn{3}{c|}{status(\%)$\uparrow$} & \multicolumn{3}{c|}{comparison(\%)$\uparrow$} & \multirow{2}{*}{acc(\%)$\uparrow$} \\
                                 &                                 & h0         & h1         & all       & h0         & h1         & all       & h0         & h1         & all        & h0         & h1         & all        & h0           & h1          & all         &                               \\ \hline
LLaVA-D & Depth   &   38.9     &   51.9     &   45.8    &   7.7     &  7.6   &  7.7 &    10.5    &   7.4   &   7.8   &   7.0    &   9.9    &   9.0    &    64.5     &    50.8    &    52.1   &   26.2  \\
LLaVA   & Camera   &   74.8     &   72.9     &   73.8    &   14.9     &  14.3   &  14.6 &    57.7    &   34.5   &   37.9   &   48.6    &   44.5    &   45.9    &    65.9     &    52.1    &    53.3   &   47.4  \\

LiDAR-LLM                        & LiDAR                           & 79.1       & 70.6       & 74.5      & 15.3       & 14.7       & 15.0      & 59.6       & 34.1       & 37.8       & \textbf{53.4 }      & 42.0       & 45.9       & 67.0         & 57.0        & 57.8        & 48.6                          \\
\textbf{Ours-LLaMA2}                          & Occ                             & 80.6       & \textbf{79.3}       & 79.9      & 18.6       & 19.1       & 18.9      & \textbf{64.9}       & 39.0       & 42.8       & 48.0       & \textbf{49.6}       &\textbf{ 49.1}       & \textbf{80.6}         & 63.7        & 65.2        & 53.4                          \\
\textbf{Ours-LLaMA3.1 }                         & Occ                             & \textbf{82.9}       & 79.2       & \textbf{80.9}      & \textbf{19.2}       & \textbf{19.2 }      & \textbf{19.2}      & 64.8       & \textbf{43.1}       & \textbf{46.3}       & 51.0       & 46.1       & 47.8       & 76.5         & \textbf{65.6 }       & \textbf{66.6}        & \textbf{54.5}                          \\ \hline
\end{tabular}
\caption{Quantitative results on NuScenes-QA. LLaVA-D input depth generated from point cloud.}
\label{tab:nuscenesqa}
\end{table*}

\subsubsection{Motion Planning} 

As illustrated in \Cref{tab:planning}, We compare the motion planning capabilities of OccLLaMA with several strong baselines that utilize various inputs and supervisions. We also compare our model with OccWorld under different settings as those in 4D occupancy forecasting task. We observe that UniAD~\cite{hu2023planning} delivers the best performance, while the supervision annotations, limiting its scalability to large-scale datasets. As an alternative, OccLLaMA achieves competitive performance relying solely on 3D semantic occupancy, demonstrating its potential to scale as a foundational model in autonomous driving. Compared to methods using occupancy as input, OccLLaMA significantly outperforms OccNet~\cite{tong2023scene}, further highlighting the superiority of the Autoregression. Additionally, surpassing the autoregressive state-of-the-art method OccWorld demonstrates the effectiveness of our pipeline. Moreover, the non-trivial performance achieved by integrating existing methods showcases the generalizability of our approach. Notably, outputting trajectories without alternating scene predictions results in performance drop, suggests that world model paradigm holds greater potential.

\subsubsection{Visual Question Answering}

To the best of our knowledge, we are the first MLLM to utilize Occupancy data with textual instructions as input and implement a series of 3D tasks in autonomous driving. We choose LiDAR-LLM~\cite{yang2023lidar}, the state-of-the-art on NuScenes-QA benchmark that integrates LiDAR into LLMs, as our primary baseline for comparison. Additionally, we evaluated a robust 2D LLM on the NuScenes-QA benchmark using depth images and raw images as input separately. We assess model performance using Top-1 accuracy metric and conduct separate evaluations for different question types. To ensure fairness, we implemente our pipeline under LLaMA2-7b, the same base model as LiDAR-LLM~\cite{yang2023lidar} and LLaVA~\cite{liu2024visual}.
As illustrated in \Cref{tab:nuscenesqa}, we observe OccLLaMA delivers the best performance overall. Compared to LiDAR-LLM, OccLLaMA can capture semantic information in 3D space better, which is essential for object-related questions. Additionally, OccLLaMA incorporates spatial information as input and aligns semantic and spatial data naturally, which is beneficial for questions involving spatial relationships.

\begin{table}[!h]
\centering
\begin{tabular}{ccc|cc}
\hline
\multicolumn{3}{c|}{Setting}  & \multicolumn{2}{c}{Reconstruction} \\
Res. & Dim. & Size & MIOU(\%)$\uparrow$   & IOU(\%)$\uparrow$   \\ \hline
50         & 128       & 2048 &        65.93          &       57.66          \\ 
25         & 256       & 2048 &        59.04          &       49.25          \\
50         & 256       & 1024 &         68.26        &       58.81        \\
50         & 256       & 4096 &        70.94         &        61.03         \\
50         & 256       & 2048 &        \textbf{75.20}          &       \textbf{63.76} \\
\hline
\end{tabular}
\caption{Ablation study of tokenizer parameters. Res. refers to resolution. Dim. refers to feature dimension. }
\label{tab:tokenizer_ablation}
\end{table}

\subsection{Ablation Study}
\subsubsection{Scene Tokenizer Parameters}

\Cref{tab:tokenizer_ablation} compares the impact of different hyperparameters on the reconstruction performance of the Scene Tokenizer, including latent space resolution, feature dimension, and codebook size. A larger codebook leads to overfitting and inefficient codebook utilization. A smaller codebooks and feature dimensions fail to effectively model the scene distributions. The resolution is positively correlated with reconstruction ability and has the most significant impact. However, larger resolution result in a greater token number to reconstruct a scene, thereby increasing the burden on forecasting. 

\begin{table}[!h]
\centering
\setlength{\tabcolsep}{1.5mm}
\begin{tabular}{cc|cc|cc}
\hline
\multicolumn{2}{c|}{Setting}        & \multicolumn{2}{c|}{Forecasting} & \multicolumn{2}{c}{Planning} \\
s.a. & a.t. & MIOU(\%)$\uparrow$& IOU(\%)$\uparrow$& L2(m)$\downarrow$&Coll.(\%)$\downarrow$\\
\hline
\checkmark  & \checkmark &\textbf{19.93}&\textbf{29.17}&1.14&0.49\\
\checkmark  &  &18.05&28.55&1.19&0.54\\
  & \checkmark &15.78&27.84&\textbf{1.12}&\textbf{0.48}\\  

\hline
\end{tabular}
\caption{Ablation study of model components. s.a. refers to spatial attention. a.t. refers to action tokenization.}
\label{tab:transformer_ablation}
\end{table}

\subsubsection{Generative Model Components}

We compares the impact of different components of the generative model on forecasting and planning performance. As illustrated in \Cref{tab:transformer_ablation}, w/o spatial attention means that the tokens in one scene maintain their original causal attention based on the flattened sequence order. w/o action tokenization means that waypoints are formed by concatenating tokens from original language vocabulary. We observe that using action-specific tokens, rather than relying on language vocabulary, results in performance gains on forecasting and planning. This improvement can be attributed to that action-specific tokens preserve the physical priors of waypoints while avoiding the inductive bias in language vocabulary. Additionally, we find that employing spatial attention to model spatial dependencies within the scene is essential for forecasting. However, it leads to a slight decrease in planning performance, which we attribute to spatial attention locally disrupting the global causal attention.



\begin{table}[!h]
\centering
\begin{tabular}{c|ccccc}
\hline
Setting & exist & count & object & status & comparison \\
\hline
P+F & \textbf{80.9} & \textbf{19.2} & \textbf{46.3} & \textbf{47.8} & \textbf{66.6} \\
F & 70.5 & 12.4 & 41.2 & 32.7 & 51.2 \\
\hline
\end{tabular}
\caption{Ablation study of caption pre-training. P refers pretraining. F refers fintuning.}
\label{tab:ablation_pretrain}
\end{table}

\subsubsection{Benefits of Pretraining}
As shown in \Cref{tab:ablation_pretrain}, we compare the impact of different training settings on QA performance: instruction fine-tuning from pretraining (\Cref{sec:trainstage}) versus instruction fine-tuning from scratch. We observe that pretraining for modality alignment lead to an overall improvement on VQA. This indicates that when OccLLaMA gains an understanding of basic 3D scenes and world dynamics, it can better accomplish high-level QA tasks.

\section{Conclusion}
In this paper, we propose OccLLaMA, a 3D occupancy-language-action generative world model in  autonomous driving for multitask. We introduce a novel scene tokenizer for discretization and reconstruction of occupancy scene. Furthermore, we build a unified multi-modal vocabulary that involve occupancy, language, and action modalities. Base on this vocabulary, we adapt LLM to perform the next token / scene prediction to complete multitask. Through extensive experiments on 4D occupancy forecasting, motion planning, and VQA, we demonstrate the multitask effectiveness of OccLLaMA. In the future, we will increase data diversity to further enhance the capabilities of OccLLaMA. We will also explore model quantization and distillation to address the inference delay caused by the large number of parameters.

\bibliography{ref}

\end{document}